\documentclass[runningheads]{llncs}
\usepackage[T1]{fontenc}
\usepackage{graphicx}
\usepackage{booktabs}
\usepackage{tablefootnote}
\usepackage{cite}
\usepackage{amssymb}
\usepackage{hyperref}
\usepackage{color}
\usepackage{amsmath}
\newcommand{\algoname}{\textit{FedMixStyle }}

\newcommand{\uproman}[1]{\uppercase\expandafter{\romannumeral#1}}
\DeclareMathOperator*{\argmin}{arg\,min} 
\begin{document}
%
\title{Efficient Cross-Domain Federated Learning\\ by MixStyle Approximation}
%
\titlerunning{\algoname}
%
\author{Manuel Röder\inst{1,2} \and
Leon Heller\inst{1} \and
Maximilian Münch\inst{2,3} \and
F.-M. Schleif\inst{1}
%
\authorrunning{Röder et al.}
%
\institute{Technical University of Appl. Sc. Würzburg-Schweinfurt, Würzburg, Germany \and
Center for Artificial Intelligence and Robotics, Würzburg, Germany \and
University of Groningen, Groningen, Netherlands}}
%
\maketitle              
\begin{abstract}
With the advent of interconnected and sensor-equipped edge devices, Federated Learning (FL) has gained significant attention, enabling decentralized learning while maintaining data privacy.
However, FL faces two challenges in real-world tasks: expensive data labeling and domain shift between source and target samples.
In this paper, we introduce a privacy-preserving, resource-efficient FL concept for client adaptation in hardware-constrained environments.
Our approach includes server model pre-training on source data and subsequent fine-tuning on target data via low-end clients.
The local client adaptation process is streamlined by probabilistic mixing of instance-level feature statistics approximated from source and target domain data.
The adapted parameters are transferred back to the central server and globally aggregated.
Preliminary results indicate that our method reduces computational and transmission costs while maintaining competitive performance on downstream tasks.

\end{abstract}
\section{Introduction}
McMahan et al.~\cite{pmlr-v54-mcmahan17a} introduced \emph{Federated Learning} (FL) as a solution to address data privacy and security risks as well as costs associated with traditional centralized learning.
FL allows multiple edge devices to collaborate in learning a global machine learning model under the supervision of a central server, while keeping local data on the client.
There has been significant progress in hardware and software technologies in recent years, with notable expansion in sensor-equipped edge devices.
Together with the increasing use of 5G-technology, this development has significantly pushed FL's integration in industrial applications.

\begin{figure}[!t]
\centering
	\includegraphics[width=\linewidth]{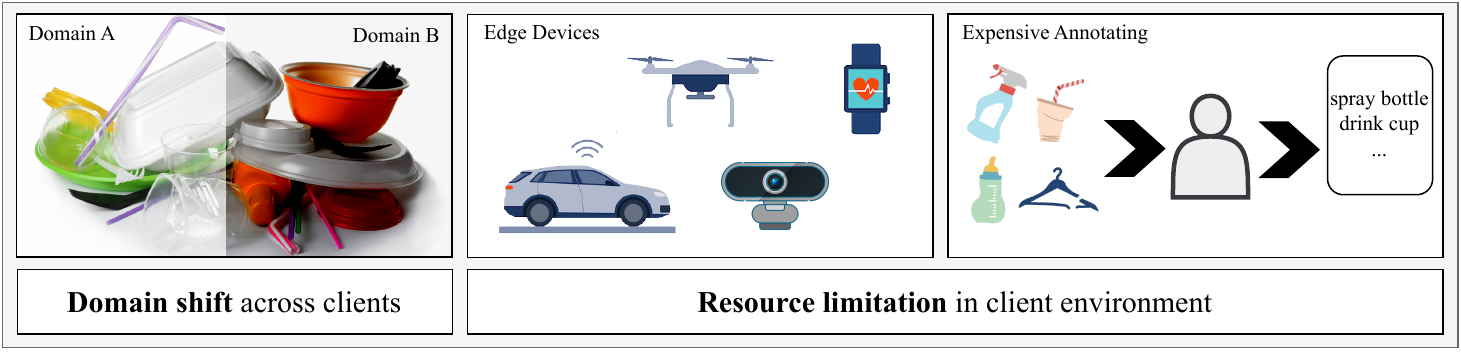}
	\caption{Challenges in federated client adaptation.
	\label{fig:problems_CSFL}
    }
\end{figure}

Nevertheless, a typical FL approach might not be appropriate in resource-restricted real-world scenarios where each client's local model is trained using isolated processing units, while data is generated locally and remains decentralized. 
Consequently, \emph{Cross-Device Federated Learning} (CDFL) \cite{10.1561/2200000083} addresses the aforementioned challenge of collaborative learning across multiple separate data silos that are limited in terms of sharing.
Moreover, the availability and quality of relevant local sensor information is influenced by a number of external factors.
Taking visual sensors as an example, domain shifts may occur as a result of variations in lighting or environmental conditions that affect contrast, color temperature, brightness, perspective, style, and noise~\cite{DBLP:journals/corr/abs-2007-01434} (see Fig. \ref{fig:problems_CSFL}).
To overcome this challenge, Zhang et al.~\cite{zhang2022lccs} proposed a BN layer modulation technique for stationary few-shot learning scenarios called LCCS.
This approach was inspired by MixStyle~\cite{zhou2021domain}, which optimizes target CNNs in classical domain generalization.

Consequently, this paper focuses on (\textbf{\uproman{1}}) integrating an established \emph{pre-train and fine-tune strategy}~\cite{chen2018a} into a CDFL setup, (\textbf{\uproman{2}}) effectively \emph{mitigating the impact of domain shift and data scarcity} during client-side fine-tune, (\textbf{\uproman{3}}) \emph{reducing computational load on clients} by adapting the BN layers of the pre-trained source model with the support samples using a re-parameterization of BN-layers and a low-dimensional approximation of the optimal target domain BN statistics as proposed in~\cite{zhang2022lccs}, (\textbf{\uproman{4}}) \emph{minimizing server-client communication overhead} whilst preserving data privacy by only transmitting target-specific BN statistic parameters back to the server instance, and (\textbf{\uproman{5}}) evaluation and \emph{aggregation of parameters at the server level} to close generalization gaps on related domains.


\section{Methodology}
In Sec.~\ref{subsec_pre} we first provide the basic mathematical notation followed by an explanation of our \algoname concept in Sec.~\ref{subsec_algo}.

\subsection{Prerequisites and Setup}\label{subsec_pre}
We consider a labeled source data set $D_S =\{X_S,Y_S\} = \{x_l,y_l\}_{l=1}^N \stackrel{i.i.d.}{\sim} p_S(X_S,$ \\$Y_S)$ in the source domain $S$ distributed by $p_S$.
For each client $i$, we have a near-target data set $D_{T_i} \stackrel{i.i.d.}{\sim} \{x_l,y_l\}_{l=1}^{k_i} \stackrel{i.i.d.}{\sim} p_{T_i}(X_{T_i},Y_{T_i})$ in the target domain $T_i$.
Samples are given as $x_l \in \mathbb{R}^d$, with $d$ indicating the number of features and $y_l \in Y$, with $Y$ denoting a discrete label space common to both source and target domain, where $|Y|=L$.
The distributions $p_S(X_S,Y_S) \neq p_{T_i}(X_{T_i},Y_{T_i})$ are subject to domain shift.
Without loss of generality, we assume for all client $i$ (with $i=1, 2, \dots, m$) that $k = k_i$ and $N \gg k$.
With $f(\phi)$ being a CNN-based deep feature extractor parameterized by $\phi$ and $g(\nu)$ denoting a nearest-centroid classifier parameterized by $\nu$, the decision function of our shared model is formalized by $\mathcal{F}(\phi, \nu) = g(\nu) \circ f(\phi)$.

\begin{figure}[!h]
\centering
	\includegraphics[width=\linewidth]{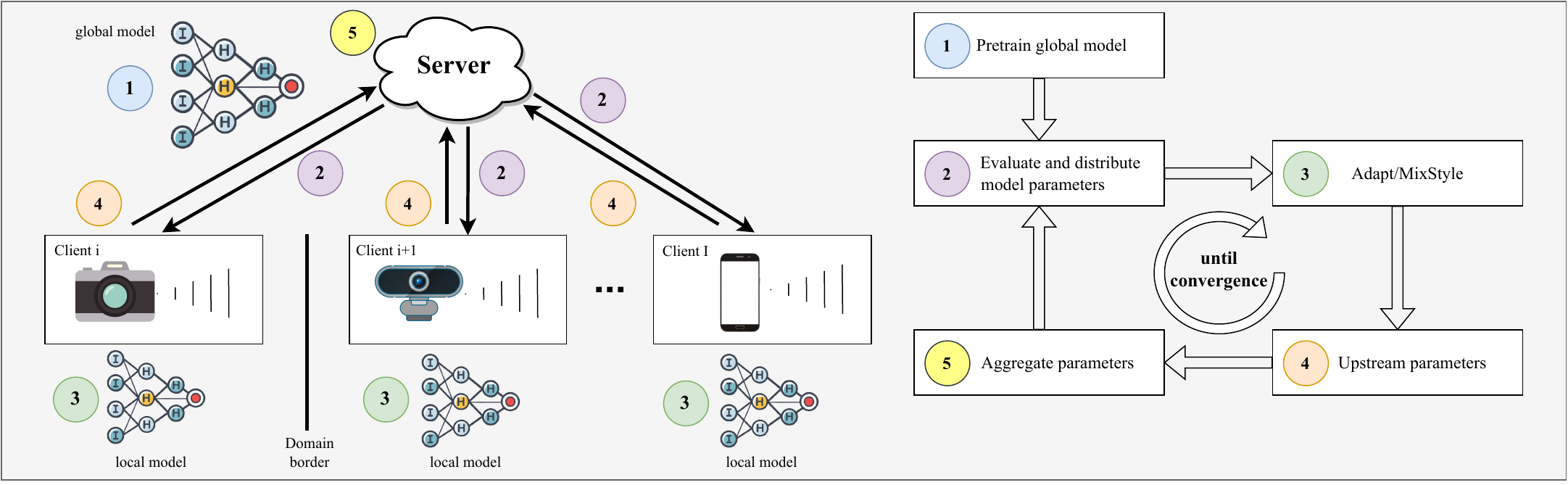}
	\caption{Conceptual overview of \algoname.
	\label{fig:cycle_Fed}
    }
\end{figure}

A central server handles the bulk of model learning, while the training data remains in separated silos on different clients with limited communication and processing power.
Given a FL setup as shown in Fig. \ref{fig:cycle_Fed}, a computationally powerful server instance can fully access the source domain data set $D_S$.
Additionally, each client $i$ participating in the distributed learning process has exclusive access to its own target domain data set $D_{T_i}$, reflecting data scarcity and privacy restrictions.
The subsequent paragraphs propose a theoretical concept that adheres to the above limitations and addresses the aforementioned challenges.
\vspace{-1cm}
\subsection{Approach}\label{subsec_algo}

Following the conceptual overview 
 of \algoname in Fig. \ref{fig:cycle_Fed}, the pre-training (\textbf{1}) objective for our server-side, supervised classification task is defined as
\begin{equation} \label{eq_srv_pretrain}
    \argmin_{\phi_S, \nu_S} \mathcal{L}_{CE}(\mathcal{F}(\phi_S,\nu_S;x_l), y_l)
\end{equation}
where $\mathcal{L}_{CE}$ denotes the cross entropy loss function and $y_l$ the ground-truth label associated with $x_l$ drawn from $D_S$ with the intention to learn and refine discriminatory and transferable features for near domains.
Consequently, the adapted parameters of the server model are used as deployment baseline for client devices joining FL (\textbf{2}) with $\phi_{T_i} = \phi_S$ and $\nu_{T_i} = \nu_S$.
The client-side adaptation of the local model (\textbf{3}) concentrates on approximating the optimal target BN statistics $\{\mu_{T_i},\sigma_{T_i}\}$ of the BN layers from the feature extractor according to~\cite{zhang2022lccs} while exploiting the BN statistics $\{\mu_{S},\sigma_{S}\}$ from the pre-trained server model and subsequently fine-tuning the learnable \emph{Linear Combination Coefficients for BN Statistics} (similar to LCCS) and target classifier by cross-entropy loss minimization on $D_{T_i}$.
In the penultimate step, clients transmit their learnt parameter set back to the server for follow-up processing (\textbf{4}). Tab.~\ref{tab:comp_transmission_size} compares typical BN methods to our approach. We show common backbone architectures and their estimated number of parameters to fine-tune as well as the expected data transmission size for one federated round, respectively.
Compared to other BN-based techniques, \algoname significantly reduces the number of client-adapted parameters.
Our method also offers the advantage of reducing the transmission sizes for server-client communication, effectively countering FL constraints.

\begin{table}[ht]
\resizebox{\linewidth}{!}{%
\begin{tabular}{lrrrrr}
\toprule
\textbf{Backbone} & \textbf{\# params} &  \begin{tabular}[c]{@{}r@{}}\textbf{\# BN}\\ \textbf{params$^*$}\end{tabular} &  \begin{tabular}[c]{@{}r@{}}\textbf{BN}\\ \textbf{size (kB)}\end{tabular} & \begin{tabular}[c]{@{}r@{}}\textbf{\# \algoname}\\ \textbf{params$^*$}\end{tabular} & \begin{tabular}[c]{@{}r@{}}\textbf{\algoname}\\ \textbf{size (kB)}\end{tabular} \\ 
\hline
ResNet-18                          & 12 million                          & 9,600                                      & 36                                                    & 80                                                                                                     & 0.3                                                          \\
ResNet-50                          & 26 million                          & 53,120                                     & 199                                                   & 212                                                                                                    & 0.8                                                          \\
ResNet-101                         & 45 million                          & 105,344                                    & 395                                                   & 416                                                                                                    & 1.6                                                          \\
DenseNet-121                       & 29 million                          & 83,648                                     & 314                                                   & 484                                                                                                    & 1.8                                                          \\ \bottomrule
\end{tabular}%
}
\caption{Client optimization parameters and transmission size comparison ($^*$ Referenced from~\cite{zhang2022lccs}, number of channels set to 1).}
\label{tab:comp_transmission_size}
\end{table}
For the sake of clarity, server-level aggregation (\textbf{5}) is carried out by applying FedAvg~\cite{pmlr-v54-mcmahan17a}. Thus, the server-side objective function, aiming to optimize generalization capabilities of our global model by parameter averaging over all client instances, cumulates to
\begin{equation}
\argmin_{\mu,\sigma} \sum_{i=1}^m \frac{\left|D_{T_i}\right|}{I} \mathcal{L}_{CE}(\mathcal{F}(\mu_{T_i},\sigma_{T_i}; x_l), y_l),
\end{equation}
where $m$ denotes the number of clients, $I$ is the total number of instances over all clients, $F$ is the shared model, and $\mathcal{L}_{CE}$ is the cross-entropy loss.
\section{Conclusion}
In this work, we presented a computationally far more efficient FL approach for solving target adaptation tasks under resource constraints and data silo distribution changes.
We plan to realize this promising concept as an extension build upon the authors' previous work (\href{https://github.com/cairo-thws/FedAcross}{https://github.com/cairo-thws/FedAcross}) and we look forward to fruitful discussions and suggestions on future work\footnote{MR and MM are supported by High Tech Agenda Bayern}.
\bibliographystyle{splncs04}
\bibliography{references_short}
\end{document}